\documentclass[conference,9pt]{IEEEtran}
\IEEEoverridecommandlockouts

\usepackage{url}
\usepackage{booktabs}
\usepackage{graphicx}
\usepackage{subcaption}
\usepackage{threeparttable}
\usepackage{multirow, multicol}

\usepackage{cite}
\usepackage{amsmath,amssymb,amsfonts}
\usepackage{algorithmic}
\usepackage{graphicx}
\usepackage{textcomp}
\usepackage{xcolor}
\usepackage[hidelinks]{hyperref}
\def\BibTeX{{\rm B\kern-.05em{\sc i\kern-.025em b}\kern-.08em
    T\kern-.1667em\lower.7ex\hbox{E}\kern-.125emX}}
\begin{document}

\newif\ifcomments
\commentstrue

\ifcomments
\newcommand{\BT}[1]{\textcolor{blue}{#1}}
\newcommand{\ML}[1]{\textcolor{purple}{ML: #1}}
\newcommand{\PB}[1]{\textcolor{red}{PB: #1}}
\newcommand{\MB}[1]{\textcolor{magenta}{MB: #1}}
\else
\newcommand{\BT}[1]{}
\newcommand{\ML}[1]{}
\newcommand{\PB}[1]{}
\newcommand{\MB}[1]{}
\fi

\title{Representation and Reference Selection in Training-Free Synthetic Image Attribution
\thanks{The corresponding author is Meiling Li (mlli20@fudan.edu.cn). This work was supported in part by the China Scholarship Council (CSC) under Grant No. 202406100170; by the Remedy project - New Frontiers (PSR 2024), funded by MUR; and by project SERICS (PE00000014) under the MUR National Recovery and Resilience Plan, funded by the European Union - NextGenerationEU.}}

\author{
Meiling Li$^{*}$, Pietro Bongini$^{\dagger}$, Benedetta Tondi$^{\dagger}$, Mauro Barni$^{\dagger}$\\
$^{*}$\textit{College of Computer Science and Artificial Intelligence, Fudan University, Shanghai, China}\\
$^{\dagger}$\textit{Department of Information Engineering and Mathematics, University of Siena, Siena, Italy}\\
}

\maketitle

\begin{abstract}

Synthetic image attribution aims at identifying the generator responsible for a given AI-generated image. Training-free reference-based attribution methods are easily scalable, since newly emerging generators can be incorporated by adding source-specific references rather than retraining a task-specific classifier. Their performance depends on two coupled factors: the representation space used for comparison and the way source-specific references are constructed. However, the interaction between these two factors remains largely unexplored. In this paper, we provide a controlled analysis of this interaction using references and off-the-shelf pretrained representations. We study representations extracted from different layers of CLIP and DINOv2, along with three reference selection strategies with varying semantic constraints: arbitrary, semantically aligned, and resynthesis-based references. Our results show that attribution accuracy generally peaks at intermediate representation levels, indicating that source-discriminative cues are more accessible before strong semantic abstraction dominates. We further show that intermediate representations are not completely semantically neutral, making reference selection critical: semantically constrained references reduce query-reference mismatch and improve attribution, especially under limited reference budgets. Resynthesis is useful in low-reference regimes, while semantically aligned references provide a better accuracy-cost trade-off when a moderate-sized reference pool is available. Our findings show that training-free reference-based attribution should be understood as the interaction between where images are compared, how the reference set is constructed, and how many references are available. Code and datasets are available at \url{https://github.com/meiling-fdu/RefPrint}.

\end{abstract}


\section{Introduction}

Identifying the generator that produced a synthetic image has become an important problem in multimedia forensics, known as \textit{synthetic image attribution} (SIA)~\cite{li2026sia}. Many existing attribution methods rely on supervised fingerprint learning, in which a classifier or feature extractor is trained to distinguish among a fixed set of known generators. Although effective in closed-set settings, such methods are difficult to scale when new generators continuously emerge, proprietary models provide only limited samples, and retraining a global attribution model is impractical\footnote{In this paper, we focus on source attribution for images assumed to be synthetic, rather than joint real-image detection and source attribution.}.

This motivates reference-based attribution, where a query image is attributed by comparing it with source-specific references rather than relying solely on a classifier trained for a fixed set of generators~\cite{cioni2024clip,bongini2025training}. Reference-based methods differ in how references are obtained and which representations are used. White-box methods exploit generator internals for inversion or reconstruction~\cite{albright2019source,zhang2020attribution,hirofumi2022did}, whereas black-box resynthesis queries candidate generators to produce query-conditioned references~\cite{bongini2025training,bongini2026source}. Under sample-only access, attribution relies on pre-collected source-labeled references without querying generators at inference time, using either off-the-shelf representations~\cite{cioni2024clip} or learned forensic representations~\cite{yang2025model,wang2026attribution,nguyen2025forensic,wu2026few}. In this work, we compare both resynthesis-based and sample-gallery-based reference selection, and study how their choices interact with off-the-shelf representations.

Reference-based attribution is governed by two coupled factors: the representation space where images are compared and the references used to characterize each candidate source. These factors are not independent, since pretrained representations usually entangle source-related artifacts with semantic information. Semantic cues are not purely a nuisance: some may be source-discriminative, as different generators may exhibit task- or content-specific priors and failure modes. However, uncontrolled semantic variation can also confound attribution by making query-reference similarity depend on content mismatch rather than source identity. In the following, we use \textit{reference selection} to broadly indicate the way the reference set is built. Its role is not to eliminate semantics, but to control query-reference semantic alignment so that useful source-related cues are preserved while accidental semantic bias is reduced.

This paper provides a controlled analysis of training-free reference-based SIA by jointly studying representation spaces, reference selection strategies, and reference budgets. For the representation space, we focus on off-the-shelf representations, without training or fine-tuning a task-specific classifier or fingerprint extractor. We analyze features extracted from different layers of pretrained vision encoders, with CLIP as the main representation and DINOv2 used to examine whether the observed trend is encoder-specific. For the reference selection, we compare three strategies with different semantic conditioning mechanisms: resynthesized references, semantically aligned references, and arbitrary references.

Our analysis leads to three main observations. First, intermediate representations generally preserve more source discrimination cues than final-layer embeddings, which emphasize higher-level semantics. Second, reference selection strongly affects performance: semantically constrained references reduce query-reference mismatch and improve attribution, especially under limited reference budgets. Third, query-specific resynthesis is useful when only one or two references are available. In contrast, semantically aligned references provide a better accuracy-cost trade-off once a moderate-sized reference pool is available. Our contributions are summarized as follows:

\begin{figure*}[t!]
    \centering
    \includegraphics[width=0.8\linewidth]{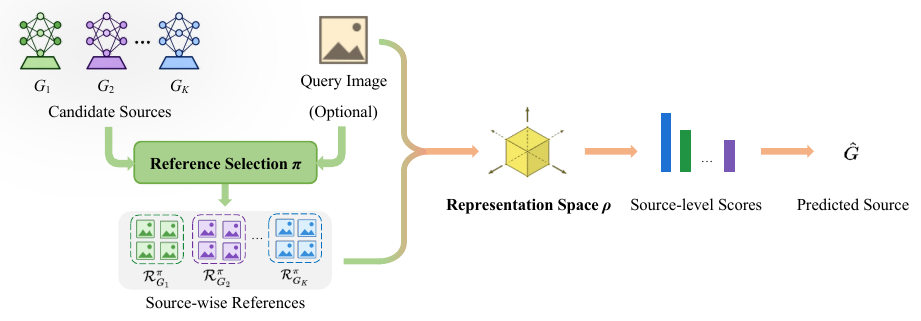}
    \caption{Reference-based attribution framework. References obtained by selection method $\pi$ are compared with the query in representation space $\rho$ and converted into source-level scores to predict $\hat{G}$.}
    \label{fig:reference_based_framework}
    \vspace{-2mm}
\end{figure*}

\begin{itemize}
    \item We provide a controlled analysis of training-free reference-based SIA by jointly investigating off-the-shelf representation spaces, reference budgets, and three reference selection strategies: resynthesis, semantic alignment, and arbitrary sampling.

    \item We construct two attribution datasets, \textit{BC-Attr-6} and \textit{COCO-Attr}, to evaluate reference selection under controlled semantic categories and diverse caption-derived content, respectively.

    \item We empirically show that intermediate-layer representations generally outperform final-layer features in the evaluated pretrained encoders, while the effectiveness of semantic alignment and resynthesis depends on the reference budget.
    
\end{itemize}

\section{Related Work}

Existing SIA methods are commonly organized into two paradigms: direct source attribution and reference-based source attribution~\cite{li2026sia}. Direct methods predict source labels from query features using a trained decision function, whereas reference-based methods compare queries with source-specific references, potentially using learned or pretrained representations.

\subsection{Direct Source Attribution}

Direct source attribution methods assume that images from the same generator share characteristic artifacts. Early approaches exploit image-domain residuals, frequency patterns, or other forensic traces introduced by generative pipelines~\cite{yu2019attributing,frank2020leveraging,goebel2021detection}. Subsequent methods learn more structured representations, including disentangled fingerprints, contrastive embeddings, and architecture-level forensic features~\cite{ding2021does,bui2022repmix,yang2022deepfake}. Recent methods further broaden the range of feature sources used for attribution. DE-FAKE~\cite{sha2023defake} combines BLIP captions~\cite{li2022blip} with CLIP visual features~\cite{radford2021learning}; MAID~\cite{zhu2025maid} exploits diffusion-model representations; OFA~\cite{fei2026one} learns transferable fingerprints through simulated architectures. These methods typically rely on learned attribution models or specialized forensic representations, which may require additional adaptation when new sources are introduced.

\subsection{Reference-based Source Attribution}

Reference-based methods attribute queries by comparing them with source-specific references, using either off-the-shelf or learned representations. These references can be obtained under different access assumptions. With white-box access, query-specific references are produced by reconstructing or inverting each candidate generator. Attribution is then obtained by selecting the generator with the highest reconstruction quality, measured in pixel, perceptual, or deep feature spaces~\cite{albright2019source,zhang2020attribution,hirofumi2022did}. With black-box access, candidate generators can be queried at inference time. Bongini et al.~\cite{bongini2025training,bongini2026source} derive a textual description from the query image, generate one reference with each candidate generator, and compare the query with these references in the final-layer CLIP feature space. Under sample-only access, candidate generators are not queried; attribution instead relies on pre-collected source samples. The $k$NN variant of Cioni et al.~\cite{cioni2024clip} matches samples using pretrained CLIP features. Other methods learn transferable forensic representations: LIDA~\cite{wang2026attribution} learns retrieval representations from low-bit-plane fingerprints, Yang et al.~\cite{yang2025model} learn model fingerprints through model synthesis, and Nguyen et al.~\cite{nguyen2025forensic} derive forensic self-descriptions for open-set attribution. Recently, OmniDFA~\cite{wu2026few} uses a learned representation for prototype-based few-shot attribution to unseen generators without test-time retraining.

These works demonstrate the effectiveness of different reference selection and representation strategies. However, their interactions across different reference budgets remain insufficiently characterized. In particular, it is unclear how query-reference semantic mismatch interacts with the source-discriminative information preserved in different representation spaces. Relatedly, Guillaro et al.~\cite{guillaro2025bias} highlight the importance of controlling semantic bias through careful dataset construction for synthetic image detection. In contrast, our work investigates semantic alignment between queries and source-specific references in SIA, jointly analyzing representation choice, reference selection, and reference budget.

\section{Reference-Based Source Attribution}
\label{sec:reference_framework}

We analyze training-free synthetic image attribution as a query-reference matching problem. Given a query image, the goal is to identify its source by comparing it with references from each candidate generator in a selected representation space. As illustrated in Fig.~\ref{fig:reference_based_framework}, this framework decomposes training-free attribution into reference selection, representation extraction, and query-reference matching. This formulation exposes three controllable factors: the reference selection method $\pi$, which determines how source-specific references are obtained, the representation space $\rho$, which determines the space where similarity is measured, and the reference budget $M$.

\begin{figure*}[t!]
    \centering
    \includegraphics[width=\linewidth]{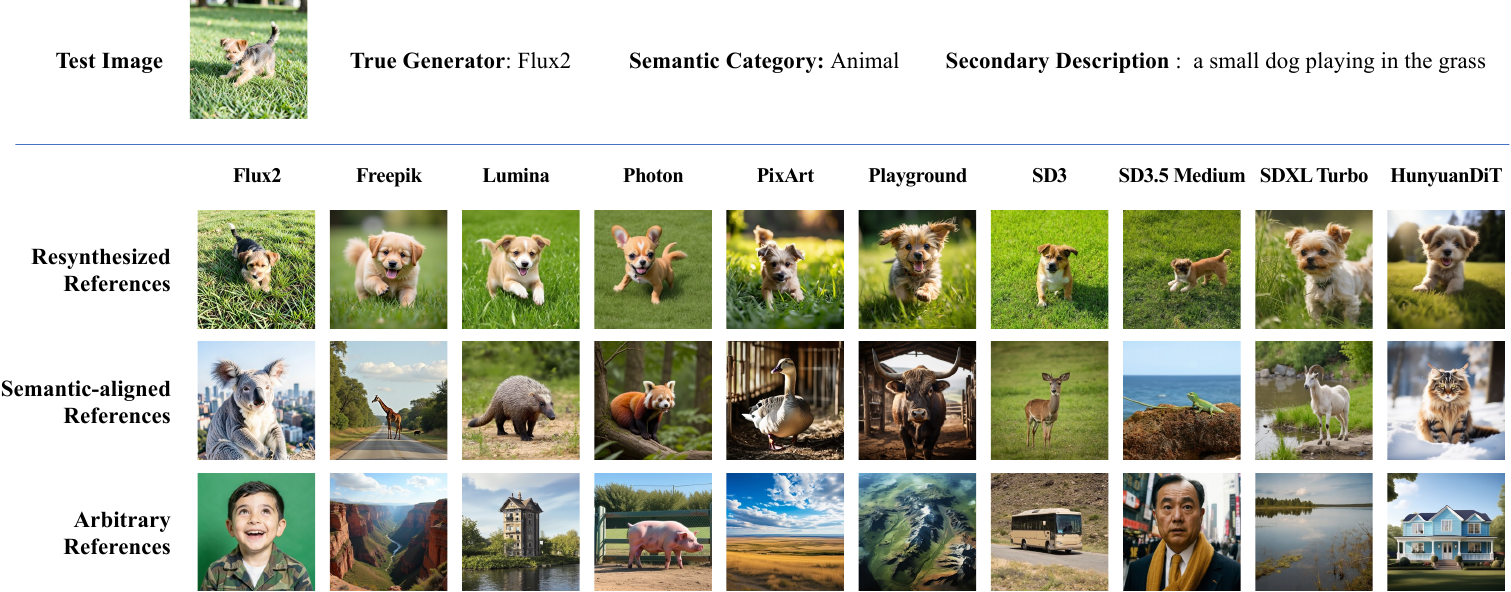}
    \caption{Reference examples on \textit{BC-Attr-6}. For a query image, resynthesized references are generated from a textual description of the query, semantically aligned references share the query's semantic category, and arbitrary references are sampled from the full source-specific pool.}
    \label{fig:ref_visual_examples}
    \vspace{-4mm}
\end{figure*}

\subsection{Unified Formulation}

Let $\mathcal{G}=\{G_1,\ldots,G_K\}$ denote the set of candidate generators. Given a query image $x_q$, closed-set attribution aims at identifying the source $G^\ast\in\mathcal{G}$ that was used to generate it. Under reference selection method $\pi$, each candidate generator $G$ is associated with a reference set
\begin{equation}
\mathcal{R}^{\pi}_{G}(x_q;M)
=
\{r^{G,\pi}_{1},\ldots,r^{G,\pi}_{M}\},
\end{equation}
where $M$ is the number of references per source. The dependence on $x_q$ indicates query-conditioned reference selection; when sampling is independent of query content, the reference set reduces to $\mathcal{R}^{\pi}_{G}(M)$, with sampling randomness left implicit.

A representation space $\rho$ specifies the feature mapping used for comparison. In our analysis, $\rho$ is instantiated by a frozen visual backbone $b$ and a selected layer $l$,
\begin{equation}
\rho=(b,l), \qquad \phi_\rho(x)=\phi_{b,l}(x).
\end{equation}

For a query-reference pair, we compute a pairwise similarity score 
\begin{equation}
a_\rho(x_q,r)
=
\mathrm{sim}\big(\phi_\rho(x_q),\phi_\rho(r)\big),
\end{equation}
where $\mathrm{sim}(\cdot,\cdot)$ is a generic similarity function.

The pairwise scores for a candidate source are aggregated into a source-level score:
\begin{equation}
    s_{\rho,\pi,M,\alpha}(x_q,G) = \alpha\big(\{a_\rho(x_q,r): r\in\mathcal{R}^{\pi}_{G}(x_q;M)\}\big),
\end{equation}
where $\alpha$ is a fixed, non-learned score aggregation rule, such as mean or maximum aggregation. Attribution is then performed as

\begin{equation}
\hat{G} = \arg\max_{G\in\mathcal{G}}s_{\rho,\pi,M,\alpha}(x_q,G).
\end{equation}
This formulation separates the effects of the representation $\rho$, the reference selection method $\pi$, the number of references $M$, and the aggregation rule $\alpha$.

\begin{table}[t!]
    \centering
    \caption{Reference selection strategies considered in this work.}
    \label{tab:reference_selection}
    \resizebox{\linewidth}{!}{%
    \begin{tabular}{lccc}
    \toprule
    Selection Method $\pi$ & Inference-time Access & Query-conditioned & Semantic Conditioning \\
    \midrule
    Resynthesis & Generator query & Yes & Caption-guided \\
    Semantically aligned & Sample pool & Yes & Category / Retrieval \\
    Arbitrary & Sample pool & No & None \\
    \bottomrule
    \end{tabular}
    }
\end{table}

\subsection{Instantiation of Reference-based Attribution}

We instantiate the framework through two design choices: the representation space and reference selection strategy, evaluated under varying reference budgets $M$.

\noindent\textbf{Representation space.}~Since attribution is based on query-reference matching, the representation defines which image properties determine the similarity score. An effective representation should preserve source-discriminative cues while reducing irrelevant semantic variation. We therefore use frozen off-the-shelf visual encoders and treat layer-wise features as representation probes. This allows us to measure how attribution performance changes across feature spaces with different abstraction levels, without assuming that a specific layer directly encodes source fingerprints.

\noindent\textbf{Reference selection.}~ The second factor is how references are obtained at inference time. We consider three strategies, summarized in Table~\ref{tab:reference_selection}. Under black-box resynthesis, candidate generators can be queried without access to their parameters or internal states. We adopt caption-guided resynthesis, where a description of the query image is used to generate references from each candidate source, requiring $KM$ generator calls per query, plus captioning. Under sample-only access, references are selected from pre-collected source-specific image pools without test-time generator calls. We consider two sample-only strategies: semantically aligned selection, which conditions references on the query's semantic content, and arbitrary sampling, which ignores query-reference semantic correspondence. Semantic alignment is defined relative to a specified semantic granularity or similarity criterion. At a coarse granularity, images may be aligned by sharing a broad category without depicting the same object or species. These strategies impose different forms of semantic conditioning, whose effects on attribution performance are empirically evaluated rather than assumed to follow a fixed ordering. 

\begin{figure*}[t!]
    \centering
    \includegraphics[width=\textwidth]{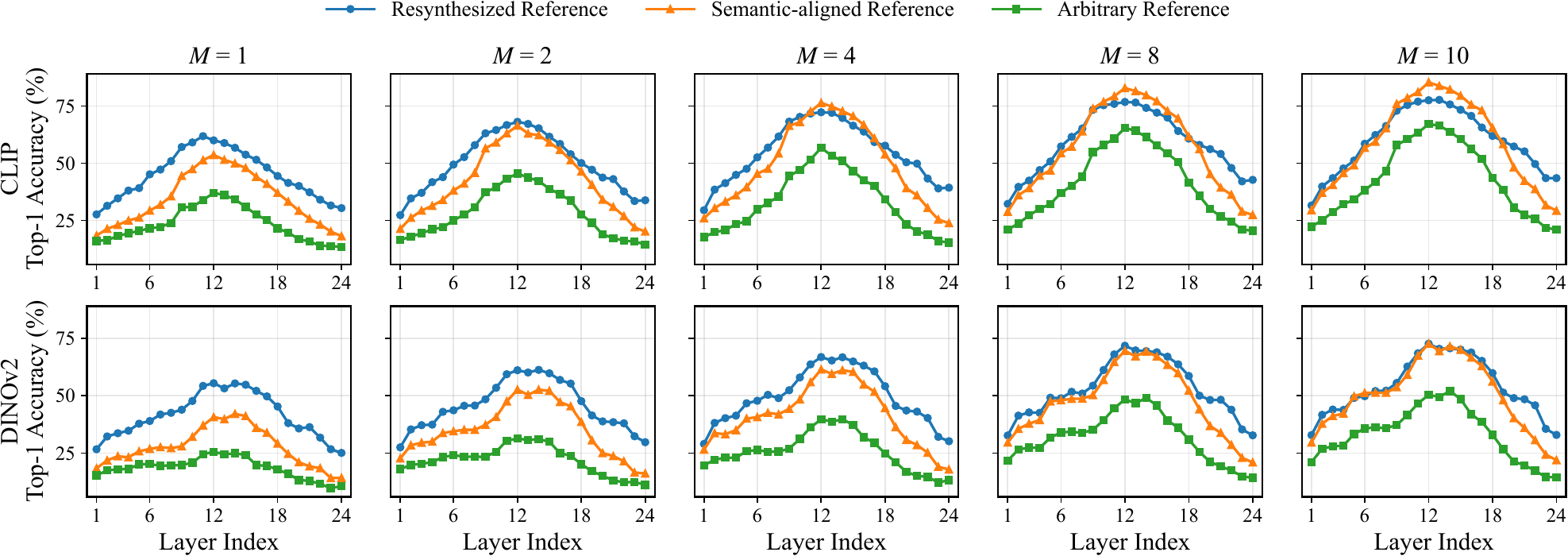}
    \caption{Attribution accuracy across CLIP and DINOv2 representation layers under different reference selection strategies and reference budgets on \textit{BC-Attr-6}.}
    \label{fig:ref_layer_analysis}
    \vspace{-4mm}
\end{figure*}

\section{Dataset Construction}

We evaluate reference-based attribution on three datasets with different levels of semantic diversity: the external face-only benchmark~\cite{bongini2025training}, denoted as \textit{FaceResyn}, and two datasets constructed for this work,
\textit{BC-Attr-6}\footnote{\href{https://huggingface.co/datasets/Meilinger00/BC-Attr-6}{https://huggingface.co/datasets/Meilinger00/BC-Attr-6}} and \textit{COCO-Attr}\footnote{\href{https://huggingface.co/datasets/Meilinger00/COCO-Attr}{https://huggingface.co/datasets/Meilinger00/COCO-Attr}}. \textit{BC-Attr-6} stands for Bias-Controlled Attribution-6: it uses balanced semantic categories and prompts shared across generators to reduce source-semantic bias. \textit{COCO-Attr} uses MSCOCO-derived captions as prompts to evaluate attribution under a more heterogeneous image distribution.

\noindent\textbf{FaceResyn Dataset.}~\textit{FaceResyn} is the 10-generator core resynthesis benchmark introduced in~\cite{bongini2025training}. It contains 1,000 head-and-shoulders portrait query images generated from 100 character descriptions. For each query and each candidate generator, the dataset provides one resynthesis produced from an alternative textual description of the same portrait, yielding 10,000 query-specific references. We use the released resyntheses as a constrained face-only setting. For the semantically aligned setting, we randomly sample one reference per source from the original image pool, excluding the query itself. Since all images are face portraits, category-level semantic alignment is inherent without explicit retrieval.

\noindent\textbf{BC-Attr-6 Dataset.}~\textit{BC-Attr-6} extends the face-only setting to controlled multi-category attribution. We use ten text-to-image generators:
$\mathcal{G}$=\{\text{Flux2}\footnote{\url{https://huggingface.co/black-forest-labs/FLUX.2-klein-4B}}, \text{Freepik}\footnote{\url{https://huggingface.co/Freepik/flux.1-lite-8B}}, \text{Lumina}\footnote{\url{https://huggingface.co/Alpha-VLLM/Lumina-Image-2.0}}, \text{Photon}\footnote{\url{https://huggingface.co/digiplay/Photon_v1}}, \text{PixArt}\footnote{\url{https://huggingface.co/PixArt-alpha/PixArt-Sigma-XL-2-1024-MS}}, \text{Playground}\footnote{\url{https://huggingface.co/playgroundai/playground-v2.5-1024px-aesthetic}}, \text{SD3}\footnote{\url{https://huggingface.co/stabilityai/stable-diffusion-3-medium-diffusers}}, \text{SD3.5 Medium}\footnote{\url{https://huggingface.co/stabilityai/stable-diffusion-3.5-medium}}, \text{SDXL Turbo}\footnote{\url{https://huggingface.co/stabilityai/sdxl-turbo}}, \text{HunyuanDiT}\footnote{\url{https://huggingface.co/Tencent-Hunyuan/HunyuanDiT-Diffusers}}\}. Prompts are grouped into six semantic categories: \{\text{faces}, \text{animals}, \text{buildings}, \text{panoramas}, \text{satellite\_views}, \text{vehicles}\}. For each generator-category pair, we generate 200 images, yielding $10 \times 6 \times 200 = 12{,}000$ images. We use 20 images per pair as queries and the remaining 180 as a pre-generated reference pool, resulting in 1,200 queries and 10,800 reference images. Semantically aligned references share the query's coarse category, whereas arbitrary references are sampled from the full source-specific pool without enforcing category matching. Fig.~\ref{fig:ref_visual_examples} illustrates these strategies on \textit{BC-Attr-6}.

\noindent\textbf{COCO-Attr Dataset.}~\textit{COCO-Attr} uses the same ten generators as \textit{BC-Attr-6}, but replaces predefined semantic categories with captions sampled from MSCOCO and used as generation prompts. This removes the fixed category structure and produces a broader distribution of objects, scenes, and compositions while keeping the candidate sources unchanged. Query and reference images are generated from disjoint MSCOCO validation and training caption splits, respectively. We use 100 query images per generator, yielding 1,000 queries and a pre-generated reference pool of 39,429 images across the ten generators. Without predefined semantic categories, this dataset also enables us to evaluate image-only semantic retrieval from the pre-generated reference pool.

\section{Experimental Results}

\noindent\textbf{Experimental setup.}~Our main experiments follow the training-free reference-based protocol described in Section~\ref{sec:reference_framework}. We compare three reference selection strategies: arbitrary sampling, semantically aligned selection, and caption-guided resynthesis. Unless otherwise specified, each candidate source is represented by $M=10$ references. For resynthesis, \textit{FaceResyn} provides one query-conditioned reference for each query-source pair, whereas for \textit{BC-Attr-6} and \textit{COCO-Attr}, we first describe the query image with LLaVA-v1.5-7B~\cite{liu2024improved} and then use this description to query each candidate generator. On \textit{COCO-Attr}, we implement image-only semantic retrieval using CLIP $L24$ to select the top-$M$ references within each candidate source. Since \textit{FaceResyn} uses ChatGPT-derived query descriptions while our datasets use LLaVA-derived descriptions, we analyze trends within each dataset rather than comparing absolute performance across datasets. For representations, we use frozen ViT-L/14 visual encoders, including CLIP-ViT-L/14@336~\cite{radford2021learning} and DINOv2-ViT-L/14 pretrained on LVD-142M~\cite{oquab2024dinov2}, both with 24 transformer layers. Each image is represented by the CLS token from the selected layer. Unless otherwise specified, we use max cosine similarity as the source-level aggregation rule. We primarily report Top-1 accuracy (ACC) under closed-set attribution, and additionally evaluate unknown-source rejection using AU-OSCR and AUROC in an open-set setting.

\begin{figure}[t]
    \centering
    \includegraphics[width=\columnwidth]{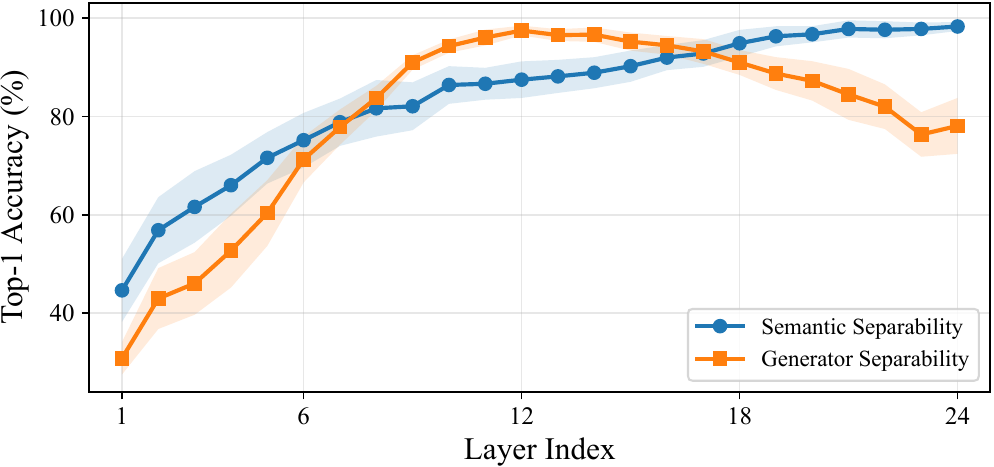}
    \caption{Semantic and generator separability across CLIP layers. We train diagnostic linear classifiers on frozen features to predict either the semantic category or the generator label. Shaded regions denote variation over the controlled factor: generators for semantic probing and semantic categories for generator probing.}
    \label{fig:separability}
    \vspace{-4mm}
\end{figure}

\begin{table*}[t!]
    \centering
    \caption{Effect of reference budget on \textit{BC-Attr-6} using CLIP $L12$. Results are mean$\pm$std (\%) over 10 independently sampled reference sets, except for resynthesis at $M=10$, which uses all available query-conditioned references.}
    \label{tab:ref_selection}
    \footnotesize
    \setlength{\tabcolsep}{4pt}
    \begin{tabular}{lccccccc}
    \toprule
    Reference Selection Method $\pi$
    & $M=1$ & $M=2$ & $M=4$ & $M=10$ & $M=25$ & $M=50$ & $M=100$ \\
    \midrule
    Arbitrary
    & $37.18\pm0.83$ & $46.23\pm1.58$ & $56.79\pm1.23$ & $68.56\pm1.35$ & $80.63\pm1.03$ & $86.20\pm0.83$ & $90.17\pm0.75$ \\
    
    Semantically aligned
    & $55.91\pm1.46$ & $66.45\pm1.08$ & $\mathbf{75.91}\pm0.67$ & $\mathbf{85.25}\pm0.98$ & $\mathbf{90.74}\pm0.83$ & $\mathbf{94.03}\pm0.60$ & $\mathbf{95.66}\pm0.29$ \\
    
    Resynthesized
    & $\mathbf{63.75}\pm1.01$ & $\mathbf{69.59}\pm0.77$ & $74.13\pm0.59$ & $77.50$ & -- & -- & -- \\
    \bottomrule
    \end{tabular}
    \vspace{-2mm}
\end{table*}

\subsection{Layer-wise Behavior on \textit{BC-Attr-6}}

Fig.~\ref{fig:ref_layer_analysis} analyzes attribution accuracy on \textit{BC-Attr-6} across CLIP and DINOv2 layers, reference selection methods, and reference budgets. Both encoders generally perform best at intermediate layers, with lower accuracy at shallow and final layers. Experiments with six CLIP variants further confirm this trend, with peak accuracy at normalized depths of 0.34--0.58. Reference selection also matters. Arbitrary references generally yield the lowest accuracy, while semantic conditioning improves attribution. Resynthesis is most effective with very few references, whereas semantically aligned selection becomes stronger as $M$ increases. Larger reference budgets generally improve accuracy without changing the preference for intermediate layers. Although CLIP outperforms DINOv2 overall, both exhibit similar layer-wise behavior.

\noindent\textbf{Separability diagnostic.}~ To interpret these results, we train diagnostic linear classifiers on frozen CLIP features to predict semantic categories and generator labels separately. These probes are used only for analysis. As shown in Fig.~\ref{fig:separability}, semantic separability increases toward deeper layers, whereas generator separability peaks at intermediate layers before declining. This suggests that intermediate representations better preserve source-discriminative information, while residual semantic information can still bias query-reference similarity, motivating semantically aligned reference selection.

\begin{table}[t!]
    \centering
    \caption{Closed-set source attribution accuracy (\%) across datasets using early ($L6$), intermediate ($L12$), and final ($L24$) CLIP representations.}
    
    \label{tab:main_results}
    \resizebox{\linewidth}{!}{
        \begin{tabular}{lccccc}
        \toprule
        Dataset & $M$ & Reference Selection Method $\pi$ & $\rho$=$L6$ & $\rho$=$L12$ & $\rho$=$L24$ \\
        \midrule
        \multirow{2}{*}{FaceResyn} & \multirow{2}{*}{1}
        & Semantically aligned & 55.20 & \textbf{89.10} & 26.70 \\
        & & Resynthesized & 70.80 & 86.70 & 57.10 \\
        \midrule
        \multirow{3}{*}{BC-Attr-6} & \multirow{3}{*}{10}
        & Arbitrary & 38.25 & 67.08 & 21.00 \\
        & & Semantically aligned & 56.67 & \textbf{85.33} & 29.25 \\
        & & Resynthesized & 58.50 & 77.50 & 43.50 \\
        \midrule
        \multirow{3}{*}{COCO-Attr} & \multirow{3}{*}{10}
        & Arbitrary & 42.50 & 75.80 & 17.70 \\
        & & Semantically aligned & 75.20 & \textbf{93.20} & 52.30 \\
        & & Resynthesized & 65.10 & 82.80 & 51.80 \\
    
        \bottomrule
        \end{tabular}
    }
\end{table}

\subsection{Cross-Dataset Analysis}

Table~\ref{tab:main_results} compares three representative CLIP layers across datasets. On \textit{FaceResyn}, random references are already category-aligned, so we omit a separate arbitrary baseline. On \textit{COCO-Attr}, which lacks predefined semantic categories, we implement image-only visual-semantic retrieval: CLIP $L24$ selects the top-$M$ references from each source's pre-generated pool, while $L6$, $L12$, or $L24$ is used for attribution. When attribution also uses $L24$, max matching is equivalent to nearest-neighbor search over the full pre-generated reference pool.

Across all datasets and reference strategies, $L12$ achieves the highest accuracy, confirming the advantage of intermediate representations. Semantic conditioning generally improves attribution, but resynthesis is not always superior: semantically aligned references outperform resynthesis on \textit{FaceResyn} ($89.10\%$ vs. $86.70\%$) and \textit{BC-Attr-6} ($85.33\%$ vs. $77.50\%$) at $L12$. On \textit{COCO-Attr}, visual-semantic retrieval achieves $93.20\%$ accuracy at $L12$, compared with $75.80\%$ for arbitrary sampling and $82.80\%$ for resynthesis. This demonstrates effective reference selection without predefined semantic labels or test-time generator access.

\subsection{Additional Analyses}

\noindent\textbf{Effect of Reference Budget.}~ We fix the representation to CLIP $L12$ and vary the number of references per source on \textit{BC-Attr-6}. Table~\ref{tab:ref_selection} reports mean$\pm$std over 10 independently sampled reference sets. Resynthesis performs best under small budgets ($M\leq2$), achieving $63.75\pm1.01\%$ and $69.59\pm0.77\%$ at $M=1$ and $M=2$, respectively. Semantically aligned selection becomes preferable from $M=4$ onward, reaching $95.66\pm0.29\%$ at $M=100$. Notably, semantic alignment with $M=4$ outperforms arbitrary sampling with $M=10$ ($75.91\%$ vs. $68.56\%$), demonstrating greater reference efficiency. These results confirm that the relative effectiveness of reference selection depends on the available reference budget. Differences from Table~\ref{tab:main_results} reflect repeated reference sampling rather than changes to the attribution protocol.

\begin{table}[t]
\centering
\caption{Effect of score aggregation rules on \textit{BC-Attr-6} using CLIP $L12$ and $M=10$. All rules use the same reference galleries.}
\label{tab:score_aggregation_rule}
\resizebox{\linewidth}{!}{
\begin{tabular}{lccc}
\toprule
Aggregation Rule & Arbitrary & Semantically aligned & Resynthesized \\
\midrule
Max    & 67.08 & 85.33 & 77.50 \\
Mean    & 54.50 & 76.00 & 72.83 \\
Softmax-weighted mean    & 66.92 & 83.83 & 75.17 \\
$k$NN majority voting ($k=5$)  & 67.17 & 84.58 & 77.83 \\
\bottomrule
\end{tabular}
}

\end{table}

\noindent\textbf{Effect of Score Aggregation Rule.}~Table~\ref{tab:score_aggregation_rule} compares source-wise max, mean, and softmax-weighted similarity with global 5-NN majority voting under identical reference sets. Softmax weighting uses $w_i\propto\exp(\tau s_i)$ with $\tau=10$, while 5-NN predicts by majority vote among the five most similar references across all sources. Across aggregation rules, semantically aligned references consistently perform best. Max similarity and 5-NN voting achieve comparable accuracy, whereas mean similarity performs worse, particularly for arbitrary references. For resynthesis, global 5-NN accuracy improves from $42.92\%$ at $L24$ to $77.83\%$ at $L12$, suggesting that the intermediate-layer advantage extends beyond source-wise max aggregation.

\begin{figure}[!t]
    \centering
    \includegraphics[width=\columnwidth]{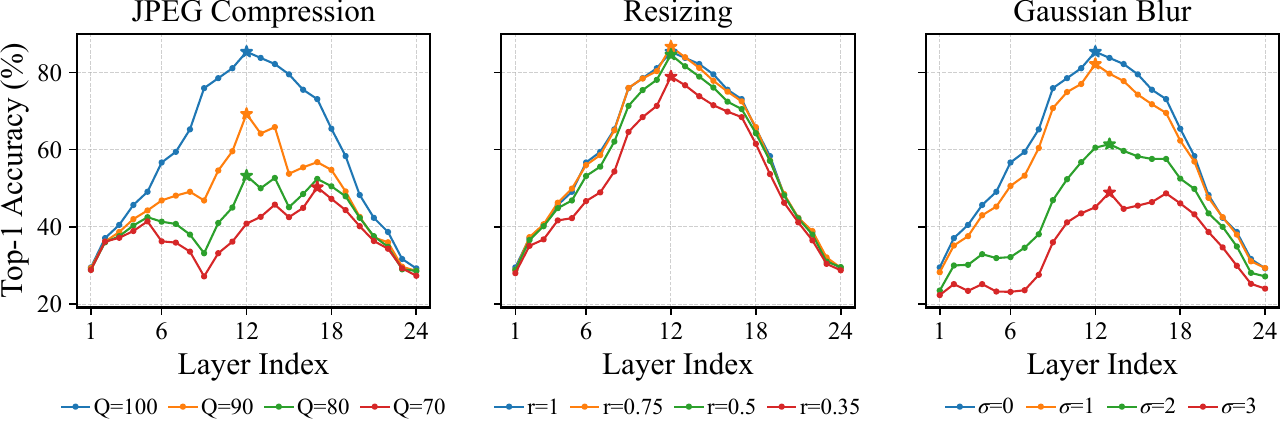}
    \caption{Attribution accuracy across CLIP layers under JPEG compression, resizing, and Gaussian blur on \textit{BC-Attr-6}, using semantically aligned references ($M=10$).}
    \label{fig:robustness_sanity}
\end{figure}

\noindent\textbf{Robustness Check.}~ Fig.~\ref{fig:robustness_sanity} shows that intermediate CLIP layers generally remain preferable under common post-processing operations, although the optimal depth can shift. Under JPEG compression at quality factor 70, accuracy peaks at $L17$ ($50.25\%$), compared with $40.83\%$ at $L12$ and $27.33\%$ at $L24$. Thus, the optimal representation depth depends on image degradation rather than remaining fixed.

\noindent\textbf{Open-Set Attribution.}~We conduct a 10-fold leave-one-generator-out evaluation, enrolling nine generators and treating the held-out generator as unknown. The maximum enrolled-source similarity serves as rejection confidence, with its threshold swept to evaluate unknown-source rejection. At $L12$ and $M=10$, known-source Top-1, AU-OSCR, and AUROC (\%) are $68.81/49.60/66.17$ (arbitrary), $86.21/65.69/73.74$ (semantically aligned), and $78.57/58.92/68.91$ (resynthesis). These results provide an initial open-set stress test, rather than comprehensive open-world attribution.

\noindent\textbf{Learned Representation Transfer.}~ We evaluate three pretrained forensic representations on \textit{BC-Attr-6} ($M=10$). Semantic alignment improves Top-1 accuracy over arbitrary references from $79.25\%$ to $86.08\%$ (Model-Fingerprint~\cite{yang2025model}), $71.67\%$ to $74.78\%$ (OmniDFA~\cite{wu2026few}), and $18.17\%$ to $22.17\%$ (LIDA~\cite{wang2026attribution}). Unlike our training-free approach, these methods require prior forensic-specific training; this experiment therefore examines the generality of reference selection rather than providing training-matched comparisons. OmniDFA's training data also overlap some evaluated generator families.

\section{Conclusion}

This paper systematically investigates training-free reference-based synthetic image attribution by jointly examining representation space, reference selection, and reference budget. Across pretrained encoders, including CLIP and DINOv2, intermediate representations generally outperform final-layer features, although the optimal depth depends on the encoder and image degradation. Semantic conditioning further improves attribution, with resynthesis being effective under very small reference budgets and semantically aligned selection becoming preferable as the budget increases. Image-only semantic retrieval demonstrates that such alignment can be achieved without predefined semantic categories or test-time generator access. Preliminary open-set results also highlight the remaining challenges of unknown-source rejection. Overall, these findings provide practical guidance for training-free reference-based source attribution.

\bibliographystyle{plain}
\bibliography{reference.bib}

\end{document}